\documentclass{article}

\PassOptionsToPackage{numbers, compress}{natbib}

\usepackage{multirow}
\usepackage{xcolor,colortbl} 
\usepackage{wrapfig}

\usepackage[preprint]{neurips_2025}

\usepackage[utf8]{inputenc} 
\usepackage[T1]{fontenc}    
\usepackage{hyperref}       
\usepackage{url}            
\usepackage{booktabs}       
\usepackage{amsfonts}       
\usepackage{nicefrac}       
\usepackage{microtype}      
\usepackage{xcolor}         
\usepackage{graphicx}
\usepackage{changepage}
\usepackage{amssymb}  
\usepackage{pifont}   
\usepackage{amsmath}
\usepackage{amssymb}
\usepackage{mathtools}
\usepackage{amsthm}
\usepackage{adjustbox}
\usepackage{multirow}
\usepackage{algorithm}
\usepackage{algorithmic}

\title{BindEnergyCraft: Casting Protein Structure Predictors as Energy-Based Models for Binder Design}

\author{%
  Divya Nori\thanks{Broad Institute of MIT and Harvard, Cambridge, MA 02142} \\
  \texttt{divnor80@mit.edu}
  \And
  Anisha Parsan\footnotemark[1] \\
  \texttt{aparsan@mit.edu}
  \AND
  Caroline Uhler\footnotemark[1] \\
  \texttt{cuhler@mit.edu}
  \And
  Wengong Jin\footnotemark[1] \, \thanks{Khoury College of Computer Sciences, Northeastern University, Boston, MA 02115} \\
  \texttt{w.jin@northeastern.edu}
}

\begin{document}

\maketitle

\begin{abstract}

Protein binder design has been transformed by hallucination-based methods that optimize structure prediction confidence metrics, such as the interface predicted TM-score (ipTM), via backpropagation. However, these metrics do not reflect the statistical likelihood of a binder–target complex under the learned distribution and yield sparse gradients for optimization. In this work, we propose a method to extract such likelihoods from structure predictors by reinterpreting their confidence outputs as an energy-based model (EBM). By leveraging the Joint Energy-based Modeling (JEM) framework, we introduce pTMEnergy, a statistical energy function derived from predicted inter-residue error distributions. We incorporate pTMEnergy into BindEnergyCraft (BECraft), a design pipeline that maintains the same optimization framework as BindCraft but replaces ipTM with our energy-based objective. BECraft outperforms BindCraft, RFDiffusion, and ESM3 across multiple challenging targets, achieving higher \textit{in silico} binder success rates while reducing structural clashes. Furthermore, pTMEnergy establishes a new state-of-the-art in structure-based virtual screening tasks for miniprotein and RNA aptamer binders.

\end{abstract}

\newcommand{\logsumexp}{\mathrm{LogSumExp}}

\section{Introduction}

\textit{De novo} protein binder design represents a fundamental challenge in molecular engineering, with wide-ranging therapeutic and biotechnological applications \citep{chevalier2017massively, gainza2018computational, yang2025design}. Recent advances in deep learning have enabled considerable progress in computational binder design, allowing the generation of binders tailored to specific target proteins and greater exploration of sequence and structural space \citep{zambaldi2024novo, watson2023novo}. However, identifying high-affinity candidates remains a bottleneck, as it typically requires generating and virtually screening thousands of designs to recover few promising hits.

\begin{figure*}[h]
    \centering
    \includegraphics[width=\columnwidth]{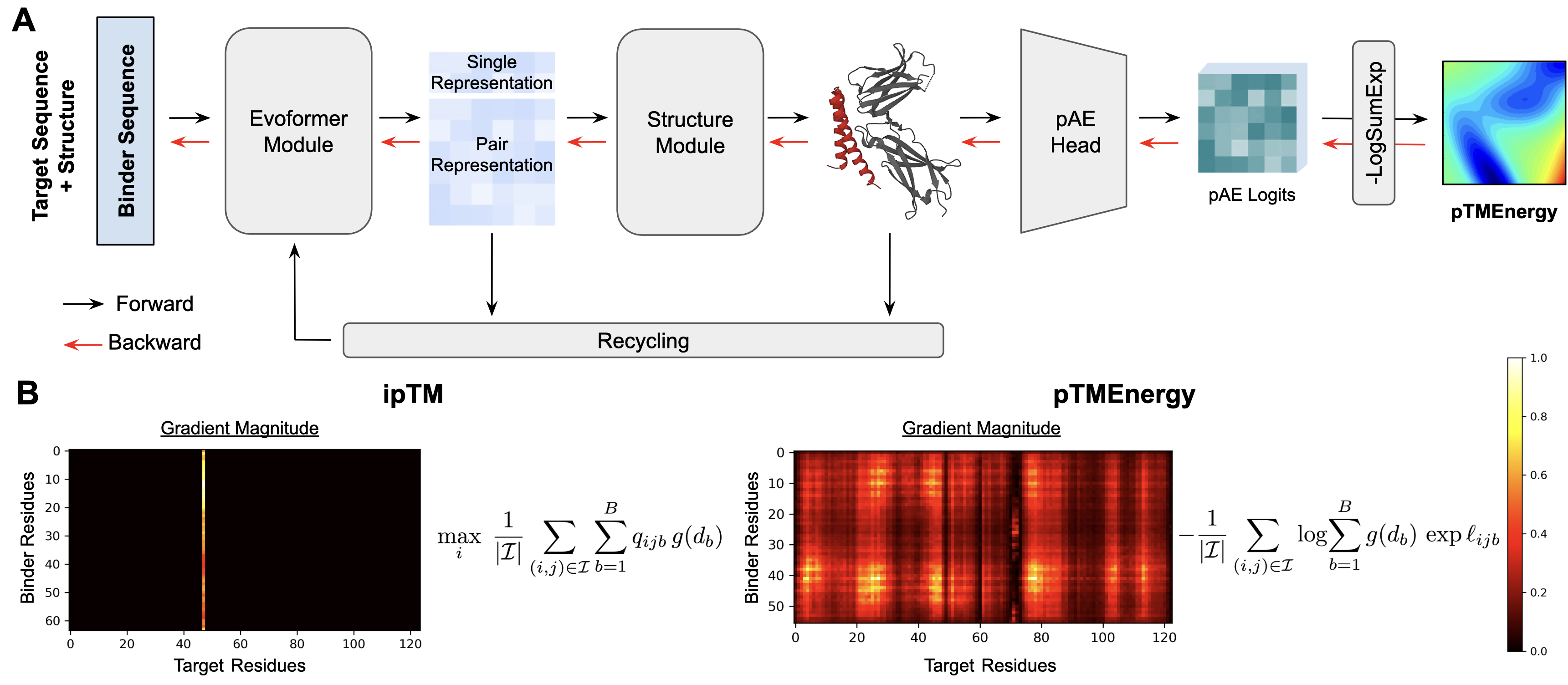}
    \caption{\textbf{A:} BindEnergyCraft (BECraft) optimizes binder sequences by backpropagating pTMEnergy, computed from pAE logits output by AlphaFold2-Multimer. \textbf{B:} Gradients from ipTM are sparse across interface residue pairs, since the maximum over target residue indices zeroes out the gradient from all but one position. pTMEnergy preserves gradients across the interface.}
    \label{fig:method}
\end{figure*}

One prominent computational design paradigm that has emerged is hallucination-based design using structure prediction models. For instance, BindCraft \citep{pacesa2024bindcraft} achieved unprecedented \textit{in vitro} success rates across multiple targets by hallucinating binder structures with AlphaFold2 and performing gradient-based optimization of the interface predicted TM-score (ipTM), among other losses, as a proxy for binding affinity. However, ipTM is fundamentally limited as an optimization objective. First, this heuristic metric does not directly reflect the statistical likelihood of the full binder–target interaction, reducing its fidelity as an objective for likelihood-based design. Moreover, as shown in Figure~\ref{fig:method}B, ipTM computes a maximum over target residue indices, resulting in sparse gradients that constrain optimization to only a small subset of interface residue pairs. 

To address these limitations, we revisit the internal confidence distributions of structure predictors. Folding models like AlphaFold2 output predicted alignment error (pAE) distributions, which quantify the model’s uncertainty over inter-residue distances. These distributions encode rich structural priors that are typically compressed into heuristics like ipTM. Inspired by the Joint Energy-based Modeling (JEM) framework \citep{grathwohl2019your}, which interprets classifier logits as unnormalized energies, we propose to reinterpret AlphaFold's pAE logits as an energy-based model over binder–target complexes. The resulting function, which we call pTMEnergy, provides a dense, differentiable signal that reflects the likelihood of a folded complex under AlphaFold’s learned distribution.

pTMEnergy offers three key advantages. First, it has a principled probabilistic interpretation, grounding hallucination-based design in binder likelihoods. Second, unlike ipTM, pTMEnergy produces dense gradients across the interface, creating a more informative and effective optimization landscape (Figure~\ref{fig:method}B). Third, it is computationally efficient: pTMEnergy is derived from the same pAE outputs used to compute ipTM, requiring no additional model calls or architecture modifications.

We incorporate pTMEnergy into the BindCraft pipeline to create BindEnergyCraft (BECraft), a hallucination framework that directly optimizes this energy-based objective (Figure~\ref{fig:method}A). Across a broad benchmark of protein targets, BECraft consistently achieves higher \textit{in silico} design success rates than BindCraft, RFdiffusion, and ESM3. We additionally show that pTMEnergy establishes new state-of-the-art performance as an unsupervised scoring function for virtual screening of both miniprotein and RNA aptamer binders, demonstrating its generality beyond hallucination-based optimization.

\section{Related Work}

\textbf{Binder Design Methods.} Recent generative methods have enabled programmable design of protein binders. Structure-based approaches such as RFDiffusion \citep{watson2023novo} and AlphaProteo \citep{zambaldi2024novo} can generate binder backbones conditioned on a target structure with experimental success. However, AlphaProteo is not an open-source method. There have been several methods developed for the design of peptides \citep{chen2024pepmlm, lin2024ppflow, li2024full}, but these do not translate well to the design of larger proteins. Hallucination-based approaches directly optimize sequences using structure predictors as proposal models. BindCraft \citep{pacesa2024bindcraft} popularized this approach by leveraging the AlphaFold2 \citep{jumper2021highly} network, achieving unprecedented experimental rates. BoltzDesign1 \citep{cho2025boltzdesign1} extends this paradigm beyond proteins to nucleic acids and ligands using the Boltz-1 structure predictor \citep{wohlwend2024boltz}.

\textbf{Binder Scoring Methods.} Scoring functions for evaluating binders span sequence-based, structure-based, and structure prediction-based methods. Sequence models like ESM-1v \citep{meier2021language} capture mutational binding effects but underperform on general binding prediction tasks. Structure-based approaches such as DSMBind \citep{jin2023dsmbind} estimate binding likelihoods using denoising, assuming access to high-quality complex structures. In contrast, structure prediction models like AlphaFold2 and AlphaFold3 \citep{abramson2024accurate} enable binding evaluation without crystal structures. Their confidence scores, particularly interface pTM (ipTM), correlate with binding strength \citep{bennett2023improving}.
\section{Methods}

Our goal is to reinterpret protein structure predictor confidence outputs as an energy-based model (EBM) that assigns a statistical likelihood to a binder–target complex, providing a rich signal to optimize binder sequence designs. While previous methods rely on ipTM score as an objective, this score is a bounded heuristic for binding that yields sparse gradients focused on just a few residue pairs. In contrast, our proposed objective produces denser, more informative gradients. We outline a general framework to extract such an energy function $E_{\theta}(x)$ in Section~\ref{sec:ebm_casting} and describe our specific instantiation, \textbf{pTMEnergy}, in Section~\ref{sec:ptmenergy}. We then integrate this energy into a gradient-based binder hallucination pipeline that iteratively refines sequences to bind a fixed target structure (Section~\ref{sec:pipeline}).

\subsection{Casting Protein Structure Predictors as EBMs}
\label{sec:ebm_casting}

We begin by formalizing a method to extract an energy function $E_{\theta}(x)$ from any black-box protein structure predictor, where $x$ is the amino acid sequence and structure of a protein complex. Under the energy-based modeling (EBM) framework, we define a probability density over $x$ as:
\begin{equation}
    p_{\theta}(x) = \frac{\exp(-E_{\theta}(x))}{Z(\theta)}
\end{equation}
where $E_{\theta}(x)$ is the unnormalized energy and $Z(\theta)$ is the intractable partition function. To derive $E_{\theta}(x)$ from a structure predictor, we leverage the observation that any classifier implicitly defines an energy function, from the Joint Energy-based Model (JEM) framework \citep{grathwohl2019your}. Consider a model $f_\theta$ that maps input $x$ to a vector of $K$ logits, where $f_\theta(x)[y]$ denotes the score for class $y$. The predicted class probabilities are given by:
\begin{equation}
    p_\theta(y|x) = \frac{\exp(f_\theta(x)[y])}{\sum_{y'} \exp(f_\theta(x)[y'])}
\end{equation}
This classifier defines a joint energy model over $(x, y)$:
\begin{equation}
    p_\theta(x, y) = \frac{\exp(f_\theta(x)[y])}{Z(\theta)}, \quad E_\theta(x, y) = -f_\theta(x)[y]
\end{equation}
Marginalizing out $y$ yields a probability over $x$ and a corresponding energy:
\begin{equation}
    p_\theta(x) = \frac{\sum_y \exp(f_\theta(x)[y])}{Z(\theta)}, \quad E_\theta(x) = -\log \sum_y \exp(f_\theta(x)[y])
\end{equation}
Thus, any classifier can be interpreted as defining an energy function via a $\text{LogSumExp}(\cdot )$ over its logits. This framework applies naturally to modern structure predictors, whose confidence heads are classifiers that predict inter-residue alignment error. 

\subsection{pTMEnergy}
\label{sec:ptmenergy}

The predicted aligned error (pAE) head outputs logits over error bins for each residue pair $(i, j)$. Specifically, for each residue pair $(i, j)$, the model outputs logits $\ell_{ij} \in \mathbb{R}^B$ over $B$ distance bins, corresponding to different magnitudes of alignment error between residues. Let $q_{ij,b} = \text{softmax}(\ell_{ij})_b$ denote the predicted bin probability, and let $d_b$ be the center of bin $b$. Let $e_{ij} = (q_{ij,1},...,q_{ij,B})$ denote the full predicted error distribution. These predictions are used to compute a global structural confidence metric known as the predicted TM-score (pTM)  \citep{evans2021protein}, defined as:
\begin{equation}
\label{eq:ptm_def}
\text{pTM}(x)
=
\max_{i}\;
\frac{1}{N}
\sum_{j=1}^{N}
\mathbb{E} \big[g(e_{ij})\big]
=
\max_{i}\;
\frac{1}{N}
\sum_{j=1}^{N}
\sum_{b=1}^{B}
q_{ijb}\,g(d_b).
\end{equation}
where the scaling kernel
\begin{equation}
g(d_b)=\frac{1}{1+\bigl(d_b/d_0(N)\bigr)^2},
\qquad
d_0(N)=1.24\,(N-15)^{\tfrac13}-1.8
\end{equation}
up-weights the contributions of low alignment errors. This reflects the intuition that accurate local packing is crucial for protein stability. The ipTM score simply subsets this computation to pairs of residues on different chains. Instead of aggregating the logits into an ipTM score, we can instead reinterpret the pAE logits as unnormalized energy scores by applying the $\text{LogSumExp}(\cdot )$ trick:
\begin{equation}
    E_{ij}(x) = -\log \sum_{b=1}^B \exp \ell_{ijb}
\end{equation}
To add the physical grounding that proteins with accurate local packing have higher likelihood, we weight each bin by $g(d_b)$ and aggregate over all inter-chain residue pairs $\mathcal{I}$, obtaining the global
\textbf{pTMEnergy}:
\begin{equation}
\label{eq:ptmenergy}
    E_{\text{pTMEnergy}}(x)
    \;=\;
    -\frac{1}{|\mathcal{I}|}
    \sum_{(i,j)\in\mathcal{I}}
        \log\!
        \sum_{b=1}^{B}
            g(d_b)\,\exp {\ell_{ijb}}.
\end{equation}

\subsection{Binder Hallucination Pipeline}
\label{sec:pipeline}

We incorporate pTMEnergy into a gradient-based binder design pipeline, \textbf{BindEnergyCraft (BECraft)}, as detailed in Algorithm~\ref{alg:becraft}. We start with a target protein sequence, and optionally structure, and initialize the binder sequence randomly. At each step, the binder sequence (parameterized as logits $z \in \mathbb{R}^{L \times 20}$) is concatenated with the fixed target sequence and passed into AF2-Multimer. The predicted complex is processed by the confidence head to produce pAE logits $\ell_{\text{pAE}}$, from which pTMEnergy is computed. This value replaces the ipTM loss component in BindCraft directly. After computing the total loss $\mathcal{L}$, gradients of $\mathcal{L}$ are backpropagated to update $z$, guiding the binder toward sequences that are well-folded, tightly interacting, and high-likelihood under the structure predictor’s confidence model. 

The overall BECraft design loss retains the structure and weights of BindCraft. In addition to pTMEnergy which has a weight of $0.05$ as ipTM did in the original BindCraft pipeline, the loss includes binder confidence via pLDDT ($0.1$), intra-binder pAE ($0.4$), inter-chain pAE ($0.1$), residue contacts within the binder ($1.0$), residue contacts between binder and target ($1.0$), and binder radius of gyration ($0.3$). All loss weights are identical to BindCraft. 

\begin{algorithm}[h]
\caption{BindEnergyCraft (BECraft): Binder Hallucination with pTMEnergy}
\label{alg:becraft}

\textbf{Input:} Target sequence $T$, optional target structure $S$
\vspace{0.25em}

Initialize binder logits $z \in \mathbb{R}^{L \times 20}$
\vspace{0.25em}

\textbf{for} each optimization step \textbf{do}
\vspace{0.25em}

\quad $x \gets \textsc{AF2-Multimer}(T, \text{softmax}(z), S)$
\vspace{0.25em}

\quad $\ell_{\text{pAE}} \gets \textsc{ConfidenceHead}(x)$
\vspace{0.25em}

\quad $E \gets \textsc{pTMEnergy}(\ell_{\text{pAE}})$
\vspace{0.25em}

\quad $\mathcal{L} \gets 0.05 \cdot E + 0.1 \cdot (1 - \text{pLDDT}) + 0.1 \cdot \text{ipAE} + 0.4 \cdot \text{pAE} - \text{con}_{\text{inter}} - \text{con}_{\text{intra}} + 0.3 \cdot \text{rad}_{\text{gyr}}$
\vspace{0.25em}

\quad $z \gets z - \eta \cdot \nabla_z \mathcal{L}$
\vspace{0.25em}

\textbf{end for}
\vspace{0.25em}

\textbf{return} $\arg\max \, \text{softmax}(z)$
\end{algorithm}

As in BindCraft, optimization proceeds through four stages. First, logit-space gradient descent explores the sequence space using softmax-relaxed amino acid distributions. At this stage, trajectories with low AF2 pLDDT are terminated. Next, a temperature annealing schedule sharpens the softmax to promote convergence toward discrete sequences. In the third stage, a straight-through estimator enables discrete sequence prediction while preserving gradient flow. Finally, a greedy mutation stage performs discrete refinement by accepting point mutations that improve the loss. We do not use MPNN-sol to redesign the interface sequence since this step is largely intended to optimize for solubility which is not relevant for the \textit{in silico} design setting considered here.

Because pTMEnergy is computed directly from model outputs already available during structure prediction, it adds no computational overhead and remains fully differentiable, making it a seamless drop-in replacement for ipTM in binder hallucination pipelines.
\section{Experiments}

We evaluate the effectiveness of pTMEnergy across four settings. Section~\ref{sec:design_success} benchmarks binder design success using BECraft, showing improved performance across eight targets. Section~\ref{sec:clashes} demonstrates that pTMEnergy leads to binder designs with fewer atomic clashes. Section~\ref{sec:ptmenergy_gradients} analyzes gradient behavior, revealing that pTMEnergy provides broader and more informative optimization signals than ipTM. Finally, we investigate whether pTMEnergy may serve not just as an optimization objective, but as a more accurate predictor of binding. Sections~\ref{sec:miniprotein} and~\ref{sec:aptamer} evaluate pTMEnergy as a structure-based binding predictor in miniprotein and RNA aptamer screening tasks.

\subsection{BECraft Improves Binder Design Success Rate}
\label{sec:design_success}

\textbf{Task.} To assess the performance of BECraft, we design binders against eight diverse protein targets from \citet{bennett2023improving}. Additional details on each target are provided in Appendix~\ref{sec:appendix_targets}. For every target, we generate 100 binder sequences with lengths between 55 and 65 amino acids, without applying any hotspot constraints. We then evaluate the \textit{in silico} design success rate, defined as the proportion of generated sequences that satisfy a specific set of structural criteria.

\begin{table}[h!]
\begin{adjustwidth}{-.6cm}{}
    \centering
    \small
    \caption{\textit{In silico} design success rates (proportion of sequences passing each constraint set) with binomial standard-error bars.\protect\footnotemark}
    \label{tab:success_rates}
    \renewcommand{\arraystretch}{1.25}
    \begin{tabular}{lcccccccc}
        \toprule
         & \textbf{ALK} & \textbf{H3} &  \textbf{IL-2R$\mathbf{\alpha}$} & \textbf{IL-7R$\mathbf{\alpha}$} & \textbf{InsulinR} & \textbf{LTK} & \textbf{TrkA} & \textbf{VirB8}  \\
        \midrule
        \multicolumn{9}{c}{\cellcolor[gray]{0.9}\textbf{Rosetta Constraints}} \\
        RFDiffusion [Unfiltered] & $.00_{.00}$ & $.02_{.01}$ & $.00_{.00}$ & $.00_{.00}$ & $.01_{.01}$ & $.00_{.00}$ & $.00_{.00}$ & $.00_{.00}$ \\
        RFDiffusion [Filtered]   & $.00_{.00}$ & $.12_{.03}$ & $.00_{.00}$ & $.02_{.01}$ & $.07_{.03}$ & $.01_{.01}$ & $.00_{.00}$ & $.01_{.01}$ \\
        \midrule
        BindCraft [no ipTM]            & $\mathbf{.25_{.04}}$ & $.16_{.04}$ & $.39_{.05}$ & $.14_{.03}$ & $.10_{.03}$ & $.40_{.05}$ & $.23_{.04}$ & $.14_{.03}$ \\
        BindCraft [ipTM + Mean]        & $.23_{.04}$ & $.08_{.03}$ & $.43_{.05}$ & $.15_{.04}$ & $.15_{.04}$ & $.15_{.04}$ & $.36_{.05}$ & $.15_{.04}$ \\
        BindCraft                       & $.14_{.04}$ & $.16_{.04}$ & $.38_{.05}$ & $.13_{.03}$ & $.13_{.03}$ & $.43_{.05}$ & $.29_{.05}$ & $.15_{.04}$ \\
        BECraft                        & $.16_{.04}$ & $\mathbf{.32_{.05}}$ & $\mathbf{.47_{.05}}$ & $\mathbf{.17_{.04}}$ & $\mathbf{.23_{.04}}$ & $\mathbf{.46_{.05}}$ & $\mathbf{.36_{.05}}$ & $\mathbf{.17_{.04}}$ \\
        \midrule
        \multicolumn{9}{c}{\cellcolor[gray]{0.9}\textbf{Folding Model Constraints}} \\
        RFDiffusion [Unfiltered] & $.02_{.01}$ & $.03_{.01}$ & $.02_{.01}$ & $.04_{.01}$ & $.07_{.01}$ & $.04_{.01}$ & $.00_{.00}$ & $.03_{.01}$ \\
        RFDiffusion [Filtered]   & $.09_{.03}$ & $.16_{.04}$ & $.10_{.03}$ & $.19_{.04}$ & $.35_{.05}$ & $.18_{.04}$ & $.00_{.00}$ & $.15_{.04}$ \\
        \midrule
        BindCraft [no ipTM]            & $.40_{.05}$ & $.40_{.05}$ & $.81_{.04}$ & $.43_{.05}$ & $.86_{.04}$ & $.66_{.05}$ & $.86_{.04}$ & $.71_{.05}$ \\
        BindCraft [ipTM + Mean]        & $.41_{.05}$ & $.50_{.05}$ & $.80_{.04}$ & $.53_{.05}$ & $\mathbf{.92_{.03}}$ & $.72_{.04}$ & $.89_{.03}$ & $.78_{.04}$ \\
        BindCraft                       & $.48_{.05}$ & $.52_{.05}$ & $.84_{.04}$ & $.51_{.05}$ & $.87_{.03}$ & $.70_{.05}$ & $.88_{.03}$ & $.72_{.05}$ \\
        BECraft                        & $\mathbf{.50_{.05}}$ & $\mathbf{.54_{.05}}$ & $\mathbf{.89_{.03}}$ & $\mathbf{.66_{.05}}$ & $.91_{.03}$ & $\mathbf{.77_{.04}}$ & $\mathbf{.89_{.03}}$ & $\mathbf{.81_{.04}}$ \\
        \bottomrule
    \end{tabular}
\end{adjustwidth}
\end{table}

\footnotetext{ESM3 is not shown in the table because its success rates are $0$ across all targets and both constraint sets.}

We use two types of constraint sets to evaluate binder quality. The first set, which we refer to as Folding Model Constraints, is based on AlphaFold2 outputs and includes thresholds on predicted structural confidence and interface quality: pLDDT greater than 0.8, interface pTM (ipTM) greater than 0.5, global pTM greater than 0.45, and interface pAE less than 0.4. These constraints are directly from the BindCraft filter criteria. The second set, referred to as Rosetta Constraints, is computed using physical interface metrics obtained from Rosetta \cite{alford2017rosetta}. These include shape complementarity greater than 0.5, dSASA greater than 1, more than 6 interface residues, more than 2 interface hydrogen bonds, surface hydrophobicity less than 0.37, and fewer than 6 unsaturated hydrogen bonds across the interface. These are again directly taken from the BindCraft filter criteria. These constraints are further explained in Appendix~\ref{sec:appendix_constraints}.

We separate these two sets of constraints because folding model metrics may advantage methods like BindCraft and BECraft, which optimize confidence scores during design. In contrast, Rosetta-based constraints provide an independent and physically grounded assessment of interface quality that does not rely on the same model used during optimization. To ensure a fair comparison across all methods, we disable all post-design filtering for both BECraft and the BindCraft baseline.

\textbf{Baselines.} In addition to benchmarking against the original BindCraft pipeline, we benchmark against two variants of BindCraft. First, we remove ipTM from the loss function to ensure that improvements observed with pTMEnergy are due to meaningful optimization signal. Second, we evaluate a version of ipTM that replaces the maximum operation with mean, a simple approach to improve gradient sparsity. We also benchmark RFDiffusion + ProteinMPNN, a widely-used two-stage pipeline that first samples backbone structures using protein backbone generative model RFDiffusion \citep{watson2023novo} and subsequently predicts sequences using inverse folding model ProteinMPNN \citep{dauparas2022robust}. While we also evaluated against ESM3 \cite{hayes2025simulating}, we found that success rates across all targets were zero across both constraint settings. Therefore, we omit this baseline. We give full details on how we ran these baselines in ~\ref{sec:appendix_baselines}. For RFDiffusion, we evaluate two conditions per target: an unfiltered setting, where 500 sequences are sampled without any selection, and a filtered setting, where the top 100 sequences are selected based on AlphaFold2-predicted ipTM scores. This setup is intended to mirror real-world design pipelines, where candidate binders are generated and then filtered post hoc.

\textbf{Results.} As shown in Table~\ref{tab:success_rates}, BECraft consistently achieves the highest \textit{in silico} design success rates across both Rosetta and Folding Model constraints. Under the Rosetta criteria, BECraft achieves highest success rate on 7 out of 8 targets, with gains up to 16 percentage points compared to the next best method. Under the Folding Model constraints, BECraft also achieves the best success rate on 7 out of 8 targets. It is interesting to note that the mean-variant of ipTM often performs better than vanilla BindCraft on a majority of targets, indicating that gradient sparsity is indeed limiting performance. Overall, BECraft demonstrates robust performance across a diverse panel of targets, achieving consistently higher rates of success regardless of the constraint regime.

\subsection{BECraft Significantly Reduces Atomic Clashes}
\label{sec:clashes}

A common failure mode in structure-based binder design is the generation of physically implausible complexes with severe atomic clashes. To assess the physical realism of our designs, we compute the fraction of sequences that result in atomic clashes after Rosetta relaxation.

\begin{figure*}[ht]
    \centering
    \includegraphics[width=\columnwidth]{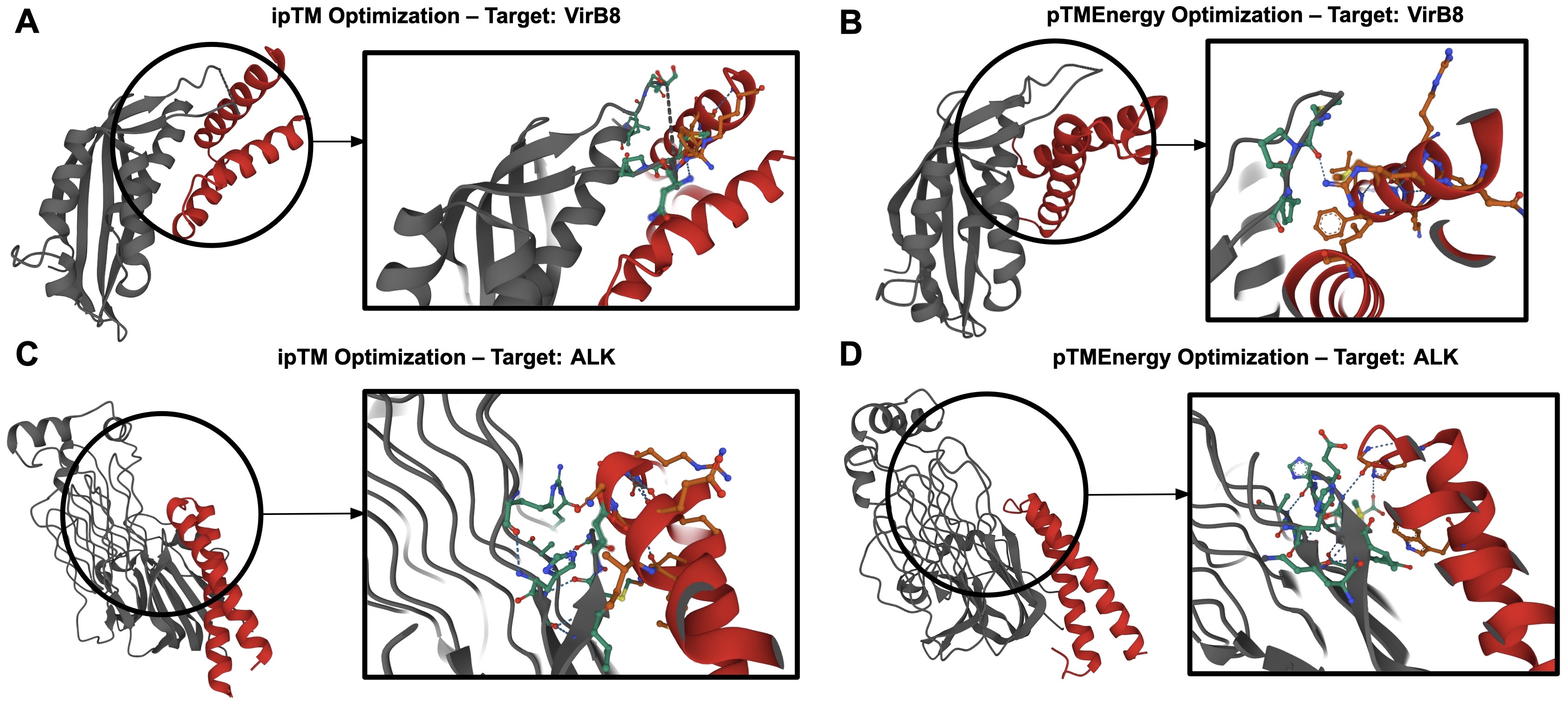}
    \caption{\textbf{(A)} A VirB8-binding complex designed using ipTM exhibits severe atomic clashes at the interface. \textbf{(B)} The corresponding pTMEnergy-optimized design for VirB8 forms a clash-free interface. \textbf{(C)} An ipTM-optimized binder for ALK with substantial steric overlap. \textbf{(D)} The pTMEnergy-optimized design for ALK forms a well-packed, physically realistic interaction.}
    \label{fig:physicality_analysis}
\end{figure*}

Table~\ref{tab:success_rates_clash} show the percentage of relaxed structures that exhibit any clashes, broken down by target. BECraft consistently produces fewer designs with clashes across all targets, reducing the clash rate from $22\%$ to $7\%$ on ALK and from $16\%$ to $5\%$ on H3.  Figure~\ref{fig:physicality_analysis} visualizes representative binder–target complexes to highlight the difference in physical plausibility between ipTM-based and pTMEnergy-based optimization. Panels A and B show designs for the VirB8 target: Panel A, optimized using ipTM, exhibits severe atomic clashes at the interface, while Panel B, optimized using pTMEnergy, shows a clean, clash-free interaction despite a similar binding mode and binder topology. A similar pattern is observed for the ALK target in Panels C and D.

\begin{table}[h!]
\begin{adjustwidth}{-1cm}{}
    \centering
    \caption{Proportion of relaxed binder–target complex structures that exhibit atomic clashes, broken down by target. BECraft reduces the clash rate compared to BindCraft across all targets.}
    \label{tab:success_rates_clash}
    \begin{tabular}{lcccccccc}
        \toprule
        & \textbf{ALK} & \textbf{H3} &  \textbf{IL-2R$\mathbf{\alpha}$} & \textbf{IL-7R$\mathbf{\alpha}$} & \textbf{InsulinR} & \textbf{LTK} & \textbf{TrkA} & \textbf{VirB8}  \\
        \midrule
        BindCraft & $.22_{.04}$ & $.16_{.04}$ & $.07_{.03}$ & $.03_{.02}$ & $.05_{.02}$ & $.03_{.02}$ & $.03_{.02}$ & $.09_{.03}$ \\
        BECraft   & $\mathbf{.07_{.03}}$ & $\mathbf{.05_{.02}}$ & $\mathbf{.04_{.02}}$ & $\mathbf{.00_{.00}}$ & $\mathbf{.04_{.02}}$ & $\mathbf{.01_{.01}}$ & $\mathbf{.01_{.01}}$ & $\mathbf{.05_{.02}}$ \\
        \bottomrule
    \end{tabular}
        \end{adjustwidth}
\end{table}

\subsection{Gradients of pTMEnergy}
\label{sec:ptmenergy_gradients}

To understand the impact of pTMEnergy compared to ipTM, we analyze the gradients each objective induces with respect to the pAE logits, which are shape $\mathbb{R}^{L \times L \times B}$ where $L$ is the total number of residues in the complex and $B$ is the number of error bins. Our goal is to assess how effectively each objective distributes learning signal across the binder–target interface during optimization.

\begin{figure*}[ht]
\centering
\includegraphics[width=\columnwidth]{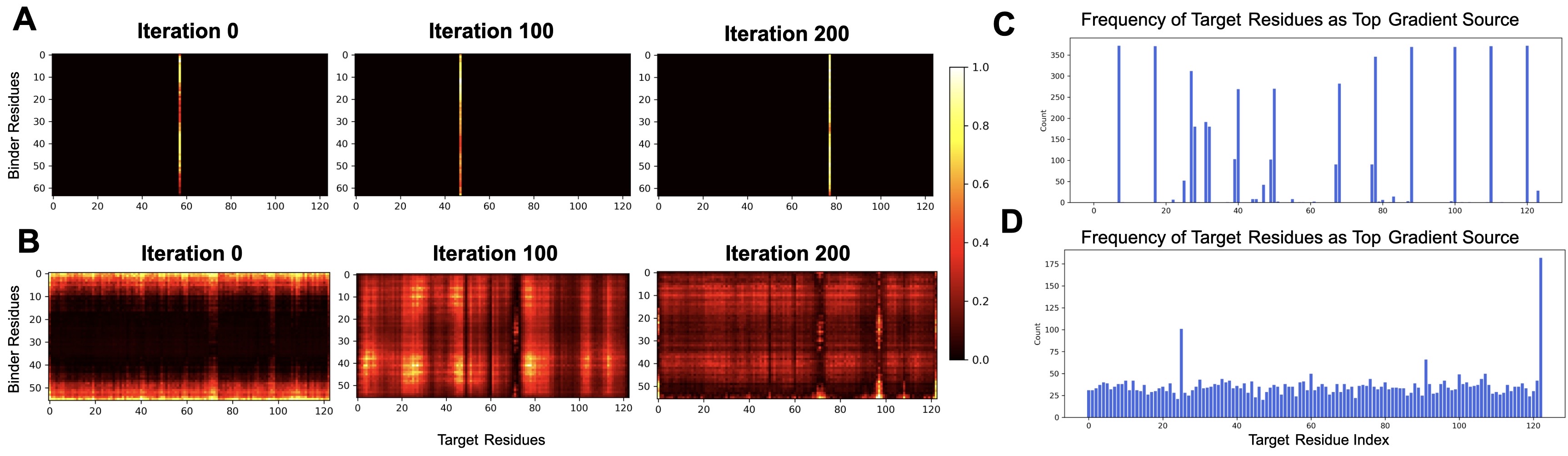}
\caption{\textbf{(A–B):} Magnitude of maximum gradient error bin for each binder–target residue pair (binder on \(y\), target on \(x\)), normalized to \([0,1]\), for ipTM and pTMEnergy respectively on an IL2Ra design task.  \textbf{(C–D):} Frequency with which each target residue ranks in the top-10 by gradient magnitude across all iterations, for ipTM and pTMEnergy respectively.}
\label{fig:gradients}
\end{figure*}

Figure~\ref{fig:gradients}A and~\ref{fig:gradients}B show the maximum gradient bin magnitude for each binder–target residue pair at steps 0, 100, and 200 of an IL2Ra design task. Gradients were computed but not used for optimization to isolate the intrinsic signal structure of each objective. Panel A corresponds to ipTM; Panel B to pTMEnergy. We find that ipTM gradients are highly localized, and each binder residue receives signal from only a single target residue per step. In contrast, pTMEnergy gradients are initially concentrated at the binder termini (step 0), but rapidly spread across the binder and target interface by steps 100 and 200.

Figure~\ref{fig:gradients}C and~\ref{fig:gradients}D quantify this difference by reporting how often each target residue appears among the top 10 contributors by gradient magnitude. Under pTMEnergy, nearly all target residues are engaged at some point, while ipTM consistently limits signal to a small subset. Across 50 independent design runs, pTMEnergy activates $98\text{--}100\%$ of target residues, compared to just $22\text{--}35\%$ for ipTM. These findings highlight pTMEnergy’s broader and more informative optimization landscape, which improves design robustness and reduces vulnerability to adversarial failure modes.

The differences in gradient behavior between ipTM and pTMEnergy arise from how each objective aggregates inter-residue alignment confidence. The ipTM score applies a hard $\max_i$ across reference residues when aggregating alignment confidence. This operation zeroes out gradients from all but the single target residue that achieves the maximum binder alignment. By contrast, pTMEnergy (Equation \ref{eq:ptmenergy}) preserves gradient signal across the full interface. Even residue pairs with moderate predicted alignment error contribute to the energy, producing broad, coordinated updates across the complex.

\subsection{Miniprotein Binder Screening}
\label{sec:miniprotein}

Having demonstrated that pTMEnergy improves binder design success and atomic clash rates, we next ask whether it can serve as an effective unsupervised predictor of binding. Traditional approaches often rely on supervised models trained to predict binding free energies, but these require large-scale, high-quality experimental data that are typically limited and expensive to obtain. Alternatively, empirical energy functions from molecular mechanics or Rosetta scoring offer physically grounded evaluations but are computationally intensive \citep{miller2012mmpbsa, schymkowitz2005foldx}. To sidestep these limitations, recent work has shown that ipTM score is a useful proxy, correlating with binding affinity in multiple contexts \citep{zambaldi2024novo}. However, these confidence metrics operate on fixed, discretized scales, limiting their effectiveness for capturing continuous, fine-grained energy landscapes. pTMEnergy, by contrast, defines a continuous energy landscape which we hypothesize makes it better suited for ranking candidate binders.

\begin{table*}[h!]
    \centering
    \caption{Virtual screening performance on miniprotein binders.}
    \vspace{0.05in}
    \begin{tabular}{llccc}
        \toprule
        & & \textbf{AUPRC} & \textbf{Precision@5} & \textbf{Precision@10} \\
        \midrule
        \multirow{4}{*}{\textbf{Baselines}}
        & Supervised Model & $0.176$ & $0.167$ & $0.167$  \\
        & Rosetta & $0.265$ & $0.375$ & $0.288$ \\
        & FoldX & $0.306$ & $0.425$ & $0.325$ \\
        & DSMBind & $0.139$ & $0.100$ & $0.088$ \\ 
        \midrule
        \multirow{3}{*}{\textbf{Boltz-1}}
        & ipTM & $0.434$ & $0.550$ & $0.525$ \\
        & ipTM + Mean & $0.455$ & $0.600$ & $0.471$ \\
        & pTMEnergy & $\mathbf{0.467}$ & $\mathbf{0.675}$ & $\mathbf{0.562}$ \\  
        \bottomrule
    \end{tabular}
    \label{tab:miniprotein}
\end{table*}

\begin{figure*}[h]
    \centering
    \includegraphics[width=\columnwidth]{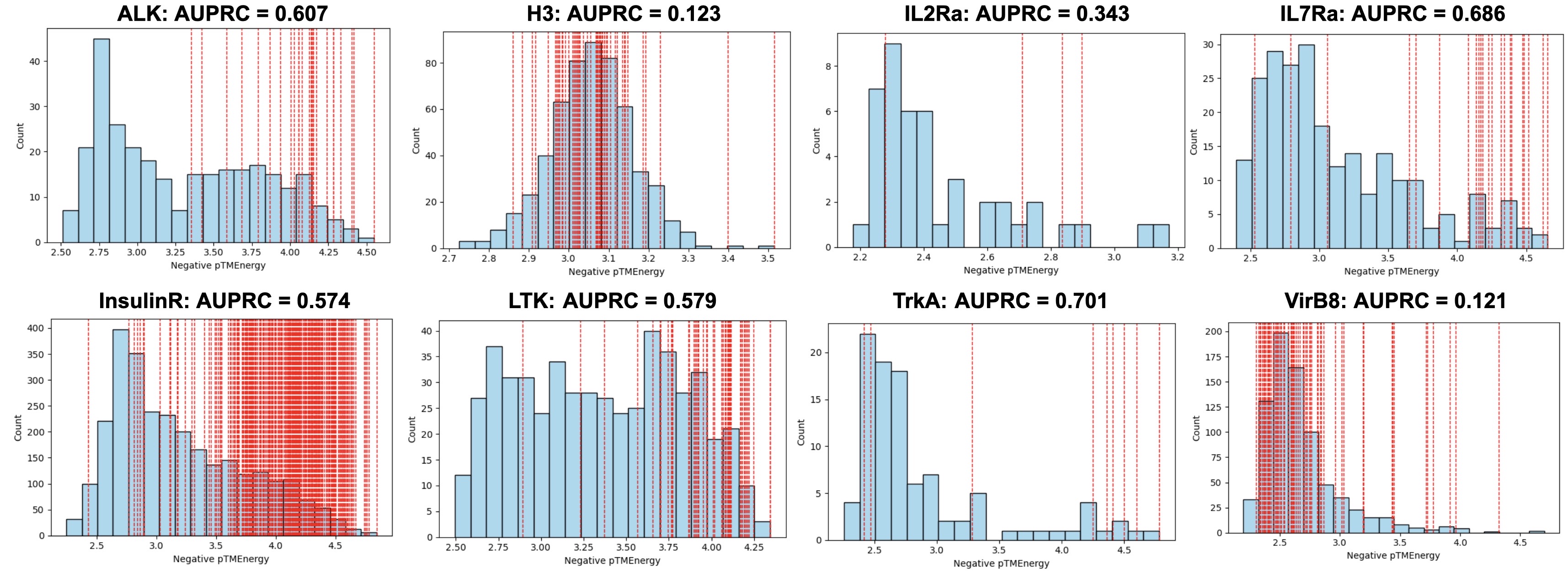}
    \caption{Distribution of predicted pTMEnergy scores. Scores are negated so that higher values indicate stronger predicted binding. Red vertical lines indicate scores assigned to true binders.}
    \label{fig:miniprotein_histograms}
\end{figure*}

\noindent\textbf{Task.} We conduct a retrospective virtual screening task using experimentally characterized miniprotein binders from \citet{bennett2023improving}. For each of the eight protein targets used in our design experiments, we evaluate whether pTMEnergy can distinguish true binders from nonbinders. Since structure prediction and MSA generation are a computational bottleneck, we subsample negative examples at a 10:1 negative-to-positive ratio. Structure prediction is performed using Boltz-1 \citep{wohlwend2024boltz}, and MSAs are computed with MMSeqs2 \citep{steinegger2017mmseqs2}.

\textbf{Baselines.} We benchmark against ipTM score as well as the modified version of ipTM where max is replaced with mean, as described in Section ~\ref{sec:design_success}. We also benchmark pTMEnergy against three state-of-the-art unsupervised binding prediction methods: two physics-based scoring functions, FoldX 
\citep{schymkowitz2005foldx} and Rosetta \citep{alford2017rosetta}, as well as DSMBind \citep{jin2023dsmbind}, a structure-based model trained to predict the likelihood of a target-binder complex as a proxy for binding. We run these baselines using the same Boltz-1 structures to ensure fair comparison, since our method does not require co-crystal structures. Additionally, to provide a comparison against a supervised method, we report the performance of model that takes ESM-2 3B \citep{lin2022language} embeddings as input and trains a feed-forward network with two hidden layers. For all 8 targets, we train on the remaining 7 and evaluate on the held-out target.

\textbf{Results.} As shown in Table~\ref{tab:miniprotein}, pTMEnergy consistently outperforms both baselines and ipTM score across all metrics. Figure~\ref{fig:miniprotein_histograms} shows the distribution of predicted energy scores (negated so that positive is predictive of binding) with red vertical lines denoting the scores of true binders. In general, true binders are assigned scores in the right tail of the distribution, though pTMEnergy does struggle for targets H3 and VirB8. These results suggest that pTMEnergy provides reliable signal for structure-based miniprotein virtual screening, even in the absence of labeled data or crystal structures.

\subsection{RNA Aptamer Binder Screening}
\label{sec:aptamer}

Finally, we assess the utility of pTMEnergy to generalize beyond protein–protein interactions to other classes of biomolecular interactions. In particular, there has been recent interest in scoring \cite{huang2024protein} and designing \cite{nori2024rnaflow} RNA aptamers that bind to a protein of interest. We evaluate whether pTMEnergy can also provide a useful signal for virtual screening of RNA aptamers.

\begin{table*}[h!]
    \centering
    \caption{Virtual screening performance on RNA aptamers binding to GFP.}

    \vspace{0.05in}
    \begin{tabular}{llccc}
        \toprule
        & & \textbf{AUPRC} & \textbf{Precision@10} & \textbf{Precision@50} \\
        \midrule
        \multirow{7}{*}{\textbf{Baselines}}
        & Transformer & $.288$ & $.233$ & $.273$ \\
        & SE(3) Transformer & $.288$ & $.200$ & $.267$ \\
        & Equiformer & $.311$ & $.300$ & $.367$ \\
        & EGNN & $.267$ & $.340$ & $.308$ \\
        & GVP-GNN & $.317$ & $.300$ & $.380$ \\
        & FA & $.290$ & $.300$ & $.333$ \\
        & FAFormer & $.322$ & $.400$ & $.413$ \\
       \midrule
        \multirow{3}{*}{\textbf{RosettaFold2NA}}
        & ipTM & $.253$ & $.150$ & $.290$ \\
        & ipTM + Mean & $.257$ & $.100$ & $.280$ \\
        & pTMEnergy & $\mathbf{.352}$ & $\mathbf{.400}$ & $\mathbf{.420}$ \\
        \bottomrule
    \end{tabular}
    \label{tab:aptamer}
\end{table*}

\textbf{Task.} Our task is to identify aptamers that bind to the Green fluorescent protein (GFP) target from a large number of screened candidates. The dataset was curated by \citet{huang2024protein}. $K_d$ values range from $0$nM to $125$nM, and aptamers with $K_d < 10$ are considered positives. We compare performance on the test set which contains $252$ positives and $686$ negatives. 

\textbf{Baselines.} We compare with all baselines reported in \citet{huang2024protein}, including methods that do not use 3D structure \cite{vaswani2017attention} and geometric models that use predicted structures \cite{liao2022equiformer, fuchs2020se, satorras2021n, jing2020learning, puny2021frame}. Structures are predicted using RosettaFold2NA (RF2NA) \cite{baek2024accurate}.

\textbf{Results.} As shown in Table~\ref{tab:aptamer}, pTMEnergy achieves the highest AUPRC among all methods, surpassing ipTM and all baselines. These findings show that pTMEnergy is effective for aptamer screening, suggesting that it may also useful in hallucination-based RNA design protocols. With the advent of structure predictors that can jointly model proteins, nucleic acids, and small molecules, it is now possible to extend hallucination-based design protocols beyond protein–protein interactions. These advances open the door to general-purpose biomolecular design frameworks that operate across molecular modalities, such as BoltzDesign1 \citep{cho2025boltzdesign1}. In the future, pTMEnergy can be integrated into this framework to serve as a potentialy beneficial optimization target for RNA–protein interaction design.

\section{Conclusions}

We introduced \textbf{pTMEnergy}, an energy function derived from protein structure predictors' inter-residue confidence outputs, and integrated it into a gradient-based hallucination framework for protein binder design. Our method, \textbf{BindEnergyCraft (BECraft)}, demonstrates improved \textit{in silico} success rates over prior approaches across diverse protein targets. Our approach also helps reduce the frequency of atomic clashes. Additionally, we showed that pTMEnergy is predictive of binding in miniprotein and RNA aptamer virtual screening tasks.

While promising, our approach has some limitations. For example, pTMEnergy inherits the failure modes of the confidence model it is based on. Hence, for interfaces where structure prediction confidence is less calibrated like antibody-antigen complexes, the binder-target likelihoods are expected to be less accurate. It is also worth noting that our current results are purely computational, and experimental validation is necessary to confirm the effects of energy-based binder design. We additionally highlight that our work raises important considerations around the responsible use of generative models for engineering novel biological functions.

Code will be released soon.

\bibliographystyle{plainnat}
\bibliography{neurips_2025}

\clearpage
\appendix

\section{Appendix}

\subsection{Protein Targets}
\label{sec:appendix_targets}

Our binder design benchmark consists of eight diverse protein targets from \citet{bennett2023improving}. Target details are given in Table~\ref{tab:miniprotein_dataset}. We subsample negative examples at a 10:1 negative-to-positive ratio.

\begin{table}[h]
    \centering
    \caption{Miniprotein dataset details.}
    \label{tab:miniprotein_dataset}
    \begin{tabular}{lccc}
        \toprule
        \textbf{Target} & \textbf{PDB ID} & \textbf{Number of Positives} \\
        \midrule
        ALK & 7NWZ & 27 \\
        H3 & 3ZTJ & 50 \\
        IL2Ra & 1Z92 & 4 \\
        IL7Ra & 3DI1 & 22 \\
        InsulinR & RFDiff SI & 259 \\
        LTK & 7NX0 & 47 \\
        TrkA & 2IFG & 9 \\
        VirB8 & 4O3V & 72 \\
        \bottomrule
    \end{tabular}
\end{table}

\subsection{Binder Design \textit{In Silico} Constraints}
\label{sec:appendix_constraints}

To assess the predicted quality of designed binder–target complexes, we evaluate two sets of \textit{in silico} constraints.

\textbf{Folding Model Constraints.} These constraints evaluate whether the designed sequence is predicted to fold correctly and form a stable, high-confidence complex with the target. All quantities are computed from AlphaFold2 outputs. They are also adapted directly from the BindCraft filtering pipeline:
\begin{itemize}
    \item \textbf{pLDDT > 0.8:} Ensures that the predicted binder structure has high backbone confidence, which is correlated with successful folding.
    \item \textbf{Interface pTM (ipTM) > 0.5:} Measures the predicted alignment between the binder and target interfaces. A higher ipTM suggests a more reliable interface geometry.
    \item \textbf{Global pTM > 0.45:} Assesses the overall complex alignment confidence. This metric encourages globally consistent binding poses.
    \item \textbf{Interface pAE < 0.4:} The average predicted aligned error at the interface must be low, indicating precise residue-residue contacts and reduced structural uncertainty at the binding interface.
\end{itemize}

\textbf{Rosetta Constraints.} These constraints evaluate the physical plausibility of the predicted complex using Rosetta \citep{alford2017rosetta}. They are also adapted directly from the BindCraft filtering pipeline:
\begin{itemize}
    \item \textbf{Shape complementarity > 0.5:} Promotes tight packing between binder and target surfaces, which is often required for strong binding.
    \item \textbf{dSASA > 1:} Ensures sufficient buried surface area at the interface, a proxy for interaction strength and stability.
    \item \textbf{> 6 interface residues:} Prevents spurious contacts by requiring that the interface involve a meaningful number of residues on the binder.
    \item \textbf{> 2 interface hydrogen bonds:} Encourages the formation of stabilizing hydrogen bonds across the interface.
    \item \textbf{Interface hydrophobicity < 0.37:} Penalizes overly hydrophobic interfaces, which are more prone to aggregation and less biologically realistic.
    \item \textbf{< 6 unsaturated interface hydrogen bonds:} Reduces the number of polar atoms at the interface that are buried but not forming hydrogen bonds. These atoms create unfavorable energy penalties, so minimizing them leads to more stable and realistic interfaces.
\end{itemize}

Together, these constraint sets enable a multi-faceted assessment of design quality, capturing both learned and physically interpretable indicators of binder success.

\subsection{Binder Design Baseline Methods}
\label{sec:appendix_baselines}

For the RFDiffusion/PMPNN baseline, we generated 500 binder backbones per target across 8 miniprotein targets using RFDiffusion \cite{watson2023novo}. We diffuse only on the binding chain from the target and do not specify hotspot residues. Binder lengths were sampled between 55–65 residues and generated with 50 diffusion steps. Each generated backbone was subsequently processed with ProteinMPNN (version \texttt{v\_48\_020}) \cite{dauparas2022robust} to design one sequence per backbone, keeping the target chain fixed and using a sampling temperature of 0.1. 

Designed sequences were then co-folded with their respective targets using AlphaFold2 \cite{jumper2021highly} and evaluated using folding model metrics (e.g., pLDDT, ipTM, interface pAE) and Rosetta-based physical constraints (e.g., shape complementarity, dSASA, hydrogen bond counts), as detailed in Section \ref{sec:design_success}. This evaluation pipeline leverages the AF2 and PyRosetta \cite{chaudhury2010pyrosetta} scheme in BindCraft.

For the ESM-3 \cite{hayes2025simulating} baseline, binder sequences were generated by prompting the model with the target chain sequence followed by a chain break token and a randomly sampled length (between 55–65 residues) of mask tokens, using 20 diffusion steps. These sequences were likewise co-folded and evaluated using the same AlphaFold2 and Rosetta-based criteria. ESM-3 was used under the EvolutionaryScale Community License Agreement, which permits non-commercial use by academic and research institutions. ProteinMPNN, RFDiffusion, and BindCraft are licensed under the MIT License. All usage in this work complies with these terms.

\subsection{Compute Requirements}

We ran all experiments on NVIDIA L40S 48GB GPUs. BECraft, and BindCraft, take approximately 5-7 minutes per binder, though it is quite dependent on complex sequence length. 

RFDiffusion takes approximately 2-3 minutes per binder generated on our dataset, though inference time varied significantly with complex sequence length. RFDiffusion reports an $O(n^2)$ scale in runtime with $n$ being sequence length \cite{watson2023novo}. Additionally, PMPNN and ESM-3 took 1-2 seconds and $\leq$ 1 minute per sequence generated respectively.

Co-folding the target and binder and scoring with AF2 designs on folding metrics took several minutes per sequence, once again varying greatly with complex sequence length. This was the most significant computational bottleneck in our scoring and evaluation of baselines.

\end{document}